\documentclass[twocolumn]{article}

\usepackage{arxiv}

\usepackage[utf8]{inputenc} % allow utf-8 input
\usepackage[T1]{fontenc}    % use 8-bit T1 fonts
\usepackage{hyperref}       % hyperlinks
\usepackage{url}            % simple URL typesetting
\usepackage{booktabs}       % professional-quality tables
\usepackage{amsmath}        % math commands (\text, etc.)
\usepackage{amsfonts}       % blackboard math symbols
\usepackage{nicefrac}       % compact symbols for 1/2, etc.
\usepackage{microtype}      % microtypography
\usepackage{graphicx}
\usepackage{natbib}
\usepackage{doi}

\title{Abstract4D: A Large-Scale Dataset and Framework for Understanding the Visual Language of Abstract Art}

\date{}

\author{
	Haowei Zhang\textsuperscript{1}, \ Yuanpei Zhao\textsuperscript{1}, \ Ji-Zhe Zhou\textsuperscript{1}, \ Mao Li\textsuperscript{1} \\
	\normalfont \textsuperscript{1}College of Computer Science, Sichuan University, Chengdu, China \\
	\normalfont \texttt{\{zhanghaowei1, zhaoyuanpei\}@stu.scu.edu.cn}, \ \texttt{\{jzzhou, limao\}@scu.edu.cn}
}

\renewcommand{\shorttitle}{Abstract4D}

\hypersetup{
pdftitle={Abstract4D: A Large-Scale Dataset and Framework for Understanding the Visual Language of Abstract Art},
pdfsubject={cs.CV, cs.AI},
pdfauthor={Haowei Zhang, Yuanpei Zhao, Ji-Zhe Zhou, Mao Li},
pdfkeywords={Abstract art, dataset, machine learning, computer vision, multimodal systems, AI-driven generation, perceptual analysis},
}

\begin{document}

% Full-width title, abstract, and keywords, with a two-column body below.
\twocolumn[
	\maketitle

	\begin{abstract}
	Artificial intelligence can classify artistic styles and synthesize images, but it still lacks a model of the visual language that gives art meaning. Abstract painting minimizes object semantics and foregrounds structural cues, making it an ideal testbed for computational perception. We introduce \textbf{Abstract4D}, the largest dataset of abstract paintings to date: more than 120,000 images paired with rich metadata and multi-dimensional prompts that capture each work's perceptual attributes---\textit{form, color, texture, and composition}. Annotations are produced by a hybrid human--VLM pipeline for quality and consistency. Using Abstract4D, we (i) analyze the semantic structure of abstract art through large-scale embedding visualization, uncovering how perceptual relationships organize artistic meaning, and (ii) establish benchmark tasks for classification, cross-modal retrieval, and text-to-image generation to evaluate how AI models perceive and reproduce abstract visual language. Together, these analyses demonstrate how Abstract4D enables both exploration and quantitative assessment of AI's ability to represent and interpret abstract art.
	\end{abstract}

	% keywords can be removed
	\vspace{0.25in}
]

\section{Introduction}
\label{sec:intro}

\begin{figure*}[!t]
	\centering
	\includegraphics[width=0.95\textwidth]{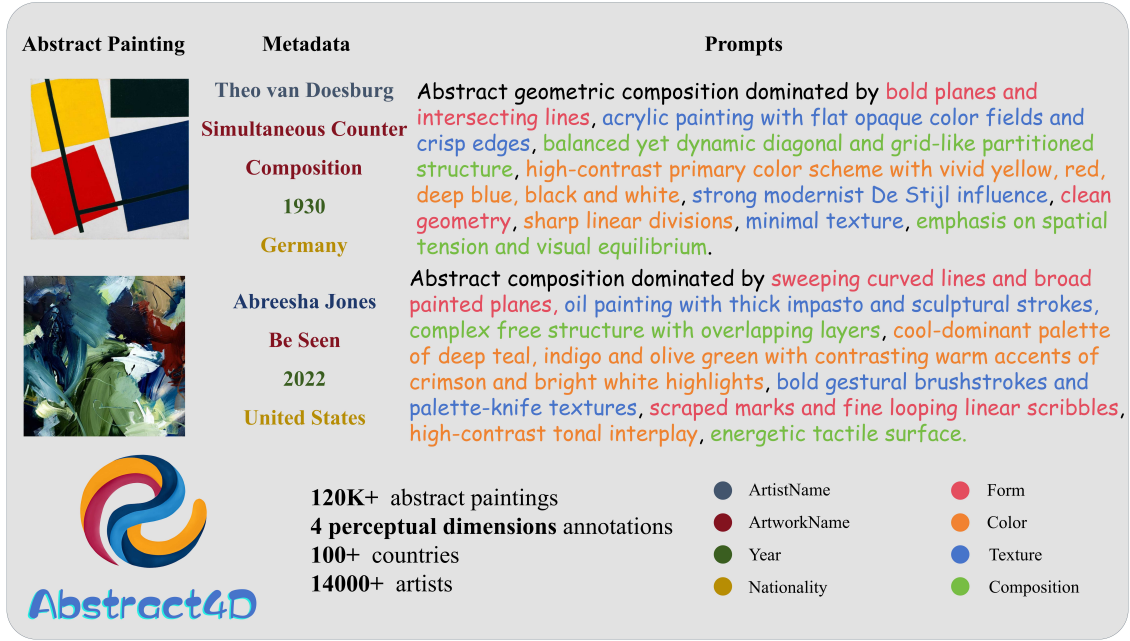}
	\caption{
		\textbf{Overview of the Abstract4D dataset.}
		Each abstract painting is paired with normalized metadata and a multi-dimensional perceptual prompt describing form, color, texture, and composition.
		Color-coded tokens indicate different perceptual dimensions.
	}
	\label{fig:abstract4d_overview}
\end{figure*}

Art has long been regarded as one of the highest forms of human cognition which is a synthesis of perception, emotion, and imagination. Unlike natural images that depict recognizable objects, artworks often convey structure and aesthetics through these visual relationships~\cite{g2022automatic}. However, despite significant progress in AI's ability to recognize objects and generate images, \textbf{understanding art} itself remains a formidable challenge.

This challenge becomes even more pronounced when dealing with \textbf{abstract art}, which deliberately distills visual elements to their most fundamental forms, often leaving behind recognizable subjects. Abstract art does not rely on traditional representation but on the \textbf{pure interplay of visual elements}, conveying meaning without the support of narrative or object-based context. The difficulty for AI in interpreting abstract art lies in this very abstraction: while AI systems excel in recognizing concrete objects, they often struggle to interpret abstract visual language~\cite{lang2021transforming, wasielewski2023computational, stork2023pixels}.

Despite increasing interest in computational aesthetics and AI-generated art, existing art datasets remain limited in capturing the internal language of art. Large-scale resources such as WikiArt~\cite{wiki2025} and OmniArt~\cite{strezoski2018omniart} have enabled style classification and metadata analysis, but they largely focus on categorical attributes rather than perceptual structure. Abstract art is also underrepresented, and most datasets lack fine-grained perceptual annotations needed to describe visual relationships such as form, composition, and texture. As a result, current AI models often rely on superficial cues such as color or texture statistics, without grasping deeper compositional or structural logic.

To address these limitations, we introduce \textbf{Abstract4D}, a large-scale dataset specifically designed to bridge the gap between perceptual understanding and computational modeling of abstract art. The dataset contains over \textbf{120,000} abstract paintings enriched with detailed metadata and \textbf{multi-dimensional annotations} that describe perceptual characteristics and structural relationships within each artwork. These annotations are organized around four fundamental perceptual dimensions---\textbf{form, color, texture, and composition}---which together constitute the core grammar of visual expression in abstract art. All annotations are generated through a \textbf{hybrid human--AI pipeline}, combining structured prompt-based VLM generation with human refinement to ensure both consistency and interpretive depth. By focusing on how these perceptual dimensions interact rather than what is depicted, Abstract4D provides a foundation for teaching AI to interpret the visual language and aesthetic structure of abstraction.

Building upon Abstract4D, we explore the perceptual organization of abstract art through two complementary approaches. First, we use \textbf{semantic embedding} and \textbf{visual analytics} to investigate the relationships between perceptual features, revealing emergent clusters corresponding to recurring artistic themes. These clusters highlight how various aspects of visual language, such as balance, expressiveness, and rhythm, manifest across different time periods and artistic movements. This approach allows us to visualize the latent structure of abstract art, illustrating how perceptual elements evolve over time and geography. As illustrated in Fig.~\ref{fig:abstract4d_overview}, Abstract4D pairs each abstract painting with structured metadata and perceptual prompts spanning four visual dimensions.

Second, we perform a series of benchmark experiments to assess how AI models perceive and internalize these perceptual structures. We evaluate state-of-the-art vision--language models on tasks including \textbf{classification} (predicting perceptual attributes from images), \textbf{cross-modal retrieval} (matching paintings with their perceptual descriptions), and \textbf{text-to-image generation} (creating abstract paintings from descriptive prompts). These experiments allow us to assess how well AI models understand the relationships among different perceptual dimensions and to identify areas where computational perception diverges from human artistic reasoning.

Together, these analyses offer a dual perspective on AI's understanding of abstract art. Through \textbf{visual analytics}, we reveal how perceptual elements organize artistic meaning within the dataset; through \textbf{benchmark evaluation}, we measure how effectively AI models capture and reconstruct this organization. By connecting data-driven analysis with artistic perception, our work bridges the gap between human aesthetic understanding and computational representation, establishing a foundation for future research on how machines can learn, interpret, and create through the visual language of art.

In summary, our contributions are as follows:

\begin{itemize}
	\item \textbf{A perceptual framework for abstract art}: We propose a structured framework that interprets abstract art as a visual language built upon the relationships among fundamental perceptual elements rather than representational content.

	\item \textbf{The Abstract4D dataset}: We release the largest and most comprehensive dataset of abstract paintings to date---over 120K images with rich metadata and multi-dimensional perceptual annotations (form, color, texture, and composition) generated through a hybrid human--AI process.

	\item \textbf{Visual analyses}: Our analyses uncover the latent organization of abstract art and identify the relationships between perceptual features, visualized through embedding and clustering techniques.

	\item \textbf{Benchmark experiments on AI models}: We establish benchmark tasks for classification, cross-modal retrieval, and text-to-image generation to assess how modern vision--language models perceive, align, and reproduce the visual language of abstract art.

\end{itemize}

\section{Related Work}

\begin{figure*}[!t]
	\centering
	\includegraphics[width=\linewidth]{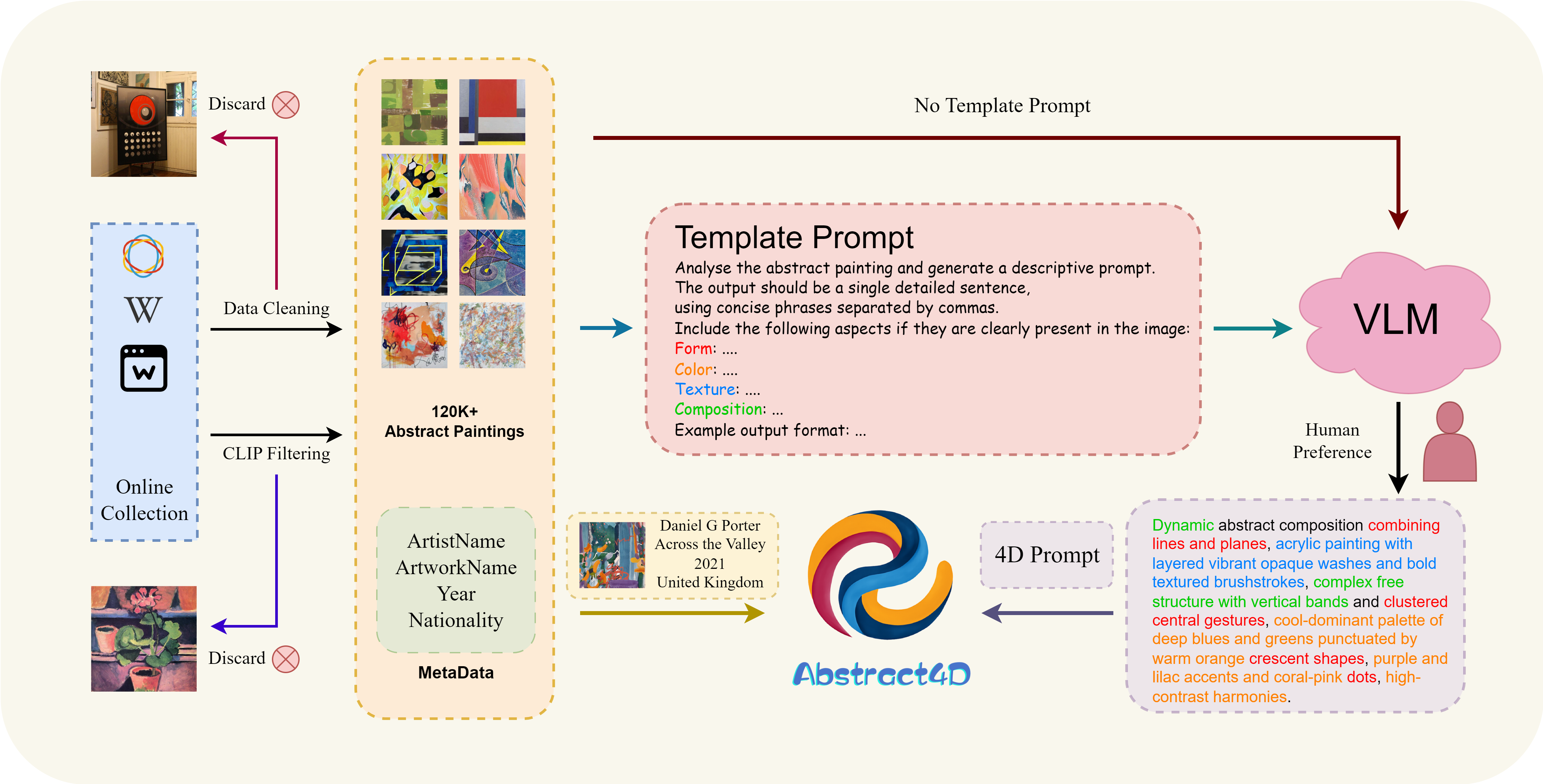}
	\caption{
		\textbf{Overview of the Abstract4D data processing and annotation pipeline.}
		Abstract artworks are collected from open sources, cleaned, and filtered through visual and semantic validation.
		Each valid painting is paired with metadata and a four-dimensional textual description generated via visual language models (VLMs)
		under both template-guided and free-form prompting.
		Human evaluators refine the outputs to form the final \textit{4D Prompt},
		integrating perceptual dimensions of form, color, texture, and composition.
	}
	\label{fig:framework}
\end{figure*}

\subsection{Foundations of Abstract Art}

Abstract art emerged in the early 20th century as a shift in how artists approached visual expression. Unlike traditional representational art, abstract art emphasizes visual language beyond physical depiction. Rather than reproducing the external world, it explores how visual elements such as form, color, and composition can convey emotional and conceptual meaning.
Wassily Kandinsky, often regarded as a pioneer of abstract art, argued that color and form could communicate spiritual and emotional states \cite{kandinsky2012concerning, duchting2000wassily, kandinsky1979point, varela2017embodied}. Piet Mondrian developed neoplasticism, reducing visual language to geometric forms and primary colors while emphasizing compositional harmony \cite{deicher1999piet, locher2005spatial}. Kazimir Malevich, through Suprematism, employed simple geometric shapes such as squares and circles to express pure artistic feeling \cite{milner1996kazimir, malevich1900cubism}.
Later artists further expanded abstraction in different directions. Mark Rothko used large color fields to evoke emotional resonance \cite{chave1989mark, breslin2012mark}, while Jackson Pollock emphasized texture and the physicality of paint to convey spontaneity and intensity \cite{solomon2001jackson}. Joan Mir\'o combined organic forms and vibrant colors to evoke subconscious imagery \cite{miro1970joan}. Today, abstract art remains a major form of visual expression, influencing numerous artistic movements worldwide.

These early artists shaped the visual language of abstract art, using form, color, texture, and composition to communicate universal emotions and intellectual experiences \cite{kress2020reading, kim2022formal}. Our framework builds on this foundation, describing abstract art through four perceptual dimensions: \textbf{form}, \textbf{color}, \textbf{texture}, and \textbf{composition}.

\subsection{Art Datasets and Representation Learning}

Existing art datasets have greatly advanced research in computational aesthetics and representation learning, yet most remain limited to categorical or stylistic annotations.
The widely used WikiArt dataset \cite{wiki2025} contains over 190,000 artworks labeled by artist and style, while OmniArt \cite{strezoski2018omniart} integrates metadata and visual features for multimodal learning.
Other large-scale collections, such as BAM! \cite{wilber2017bam} and Multitask Painting 100k \cite{bianco2019multitask}, further expand artistic content but still rely primarily on discrete style or genre categories.
More recent datasets, such as DELAUNAY \cite{gontier2022delaunay}, begin to explore abstract artworks, yet existing resources generally lack structured perceptual annotations describing how visual elements are organized within abstract paintings.

\begin{table}[t]
	\centering
	\small
	\caption{
		\textbf{Four-dimensional perceptual framework of abstract art.}
		Each dimension captures a core aspect of abstract visual perception.
	}
	\vspace{4pt}
	\resizebox{\columnwidth}{!}{%
		\begin{tabular}{l p{3.0cm} p{3.2cm}}
			\toprule
			\textbf{Dimension} & \textbf{Description} & \textbf{Representative Keywords} \\
			\midrule
			\textbf{Form} & Structural and geometric organization of visual elements. & geometric, organic, angular, curved, fragmented, fractal \\
			\textbf{Color} & Chromatic relationships of tone, contrast, and temperature. & warm, cool, monochrome, contrasting, gradient \\
			\textbf{Texture} & Perceived surface quality and material expression. & smooth, rough, layered, reflective \\
			\textbf{Composition} & Spatial arrangement and visual balance. & centered, diagonal, balanced, hierarchical, radial, dispersed \\
			\bottomrule
		\end{tabular}%
	}
	\label{tab:framework}
\end{table}

\begin{figure*}[!t]
	\centering
	\includegraphics[width=\linewidth]{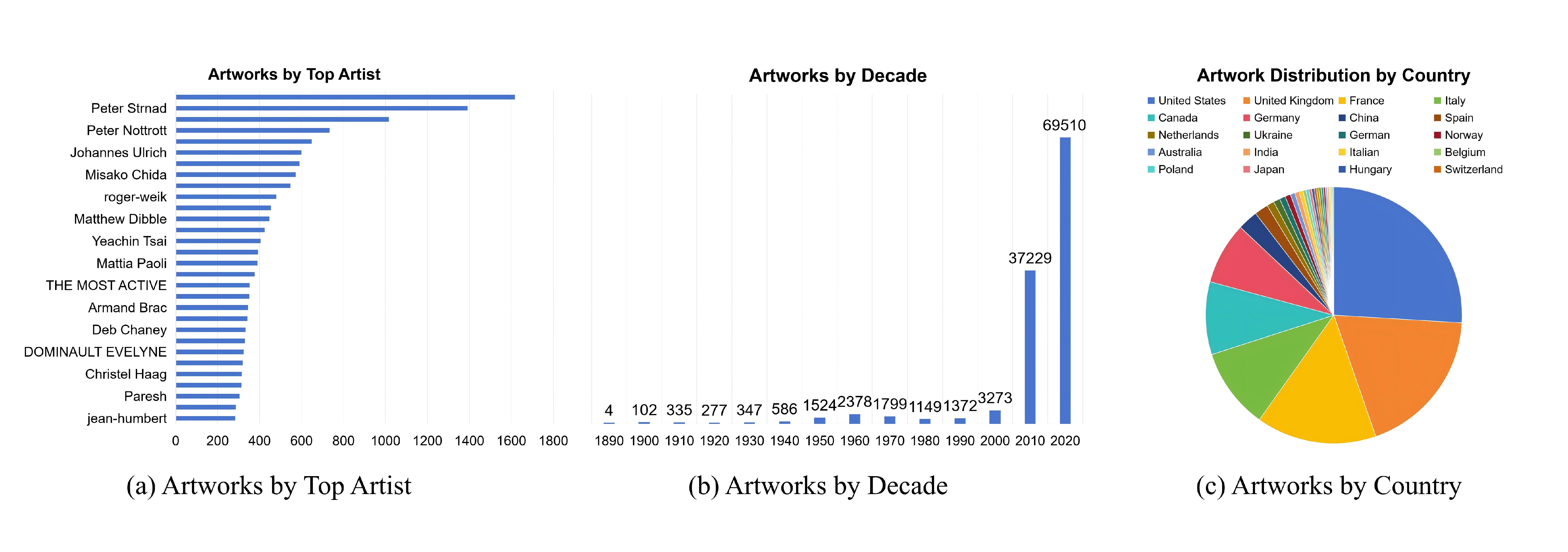}
	\caption{
		\textbf{Overview of the Abstract4D dataset statistics.}
		(a) Top-30 artists ranked by the number of artworks.
		(b) Distribution of artworks across decades, showing a marked increase after 2000.
		(c) Proportion of artworks by artist nationality.
		Together, these visualizations illustrate the temporal growth and geographical diversity
		of the Abstract4D dataset, highlighting its balance across artistic contributors,
		historical periods, and cultural origins.
	}
	\label{fig:dataset_overview}
\end{figure*}

Despite their scale, abstract artworks are underrepresented in existing datasets, which are dominated by figurative works. As a result, models trained on these datasets often focus on surface features rather than the unique visual language of abstraction. Moreover, existing datasets lack detailed perceptual annotations (e.g., form, color, texture, composition) \cite{mureika2005multifractal, elgammal2018shape}, hindering models' understanding of abstract art's structural and expressive elements.

To address these limitations, we introduce \textbf{Abstract4D}, a large-scale dataset of over 120K abstract paintings with multi-dimensional annotations that capture the visual grammar of abstract art, enabling deeper human and machine understanding.

\begin{table*}[t]
	\centering
	\small
	\caption{
		\textbf{Comparison of Abstract4D with major existing art datasets.}
		While prior datasets such as WikiArt and OmniArt provide large-scale style or metadata labels, they contain limited abstract works and lack structured perceptual annotations.
		Abstract4D uniquely focuses on \emph{purely abstract paintings}, offering multi-dimensional perceptual descriptions aligned with artistic language.
	}
	\vspace{4pt}
	\setlength{\tabcolsep}{6pt}
	\resizebox{\textwidth}{!}{%
	\begin{tabular}{lccc}
		\toprule
		\textbf{Dataset} & \textbf{\#Images} & \textbf{Abstract Focus} & \textbf{Perceptual / Semantic Annotations}  \\
		\midrule
		WikiArt~\cite{wiki2025} & 196K & $\sim$9\% (18K) abstract works & Style / genre / artist labels  \\
		BAM!~\cite{wilber2017bam} & 2.5M artworks & Low (few-abstract) & Content, media, emotion tags  \\
		OmniArt~\cite{strezoski2018omniart} & 2M & Low (few abstract) & Multi-label: style, material, technique  \\
		iMet Collection~\cite{zhang2019imet} & 130K & Minimal & Fine-grained expert attributes  \\
		ArtEmis~\cite{achlioptas2021artemis} & 80K & Low (mostly figurative) & Emotion + explanation texts  \\
		Painter by Numbers~\cite{painter-by-numbers} & 100K & Negligible & Author labels (pairs) \\
		SemArt~\cite{garcia2018read} & 23K & Moderate & Textual descriptions (semantic)  \\
		ArtBench-10~\cite{liao2022artbench} & 60
		K & $\sim$10 styles (balanced) & Style labels only \\
		\textbf{Abstract4D (ours)} & \textbf{120K} & \textbf{100\% abstract} & \textbf{Basic labels + 4 perceptual dimensions}\\
		\bottomrule
	\end{tabular}%
	}
	\label{tab:comparison}
\end{table*}

\subsection{Models' Understanding of Art}

While deep learning has advanced computational art analysis, most models still focus on surface-level visual features and struggle to capture the perceptual structures underlying abstract art. Architectures such as VGGNet \cite{simonyan2014very} and ResNet \cite{he2016deep} have been widely applied to artistic style classification and transfer \cite{gatys2015neural, johnson2016perceptual}, yet they remain limited in representing core abstract elements such as form, composition, and texture.

Recent vision--language models, including CLIP \cite{radford2021learning} and BLIP \cite{li2022blip}, align images and text in shared embedding spaces, enabling cross-modal understanding. However, their ability to interpret abstract visual language remains constrained by the lack of fine-grained perceptual annotations in existing datasets, which typically provide only style or genre labels. Abstract4D addresses this limitation by introducing structured annotations across four perceptual dimensions enabling models to better capture the visual grammar of abstract art.

\section{Dataset}
\label{sec:dataset}

The Abstract4D dataset provides a large-scale resource for studying AI perception of abstract art. It contains over \textbf{120,000} curated abstract paintings, each paired with a textual prompt describing four perceptual dimensions---\textit{form}, \textit{color}, \textit{texture}, and \textit{composition}---along with normalized metadata such as artist, nationality, and creation year.

Figure~\ref{fig:framework} presents the overall construction pipeline, which combines multimodal data collection, automatic filtering, and human--AI collaborative annotation to ensure both scale and annotation quality.

\begin{figure}[!t]
	\centering
	\includegraphics[width=0.99\linewidth]{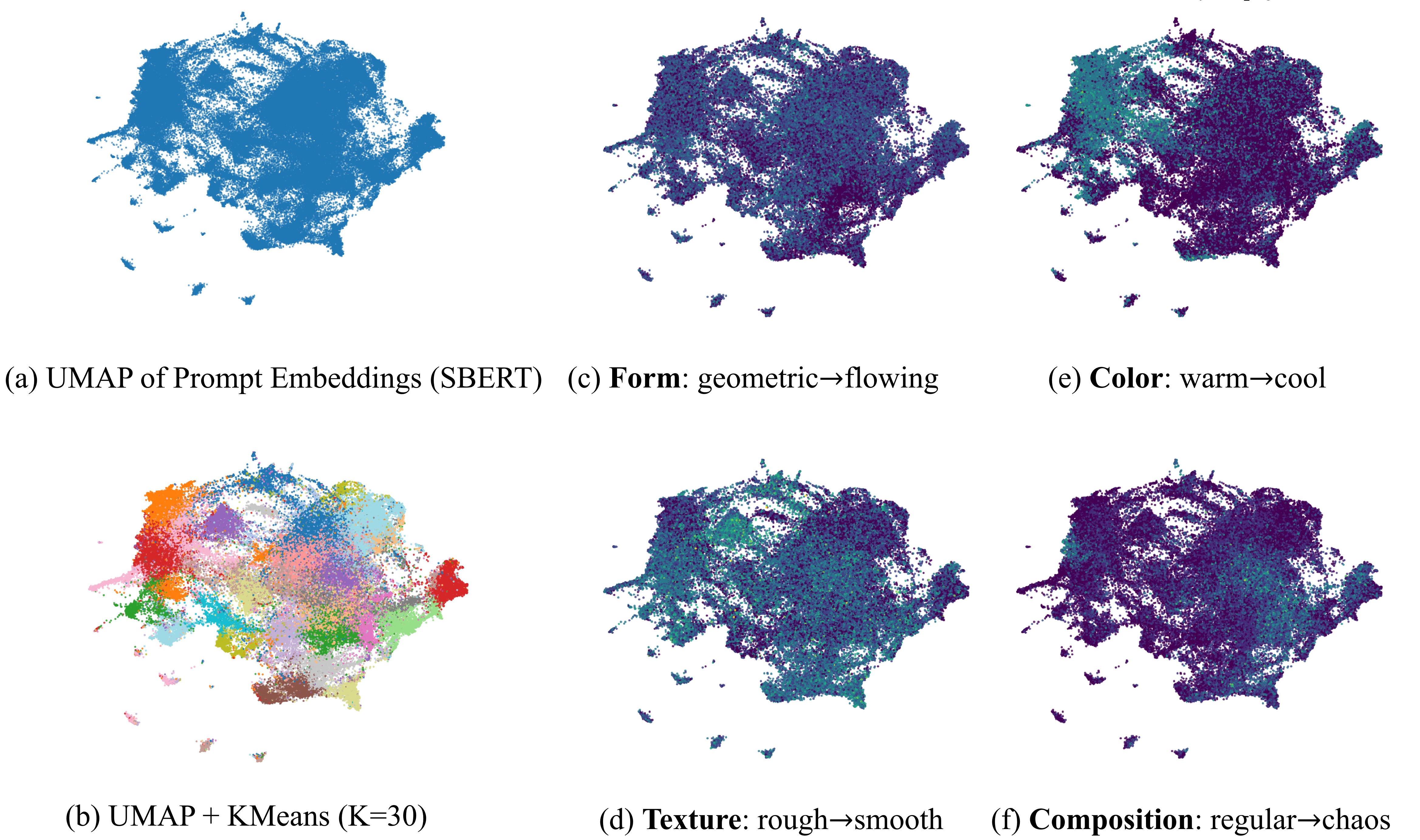}
	\caption{
		\textbf{UMAP Visualization of Abstract4D Semantic Space.}
		\textbf{(a):} Raw distribution of 120K prompts showing a continuous manifold.
		\textbf{(b):} KMeans clustering ($K{=}30$) revealing smooth transitions in perceptual themes.
		\textbf{(c)-(f):} Cross-semantic visualization color-coded by perceptual dimensions (\textit{Form}, \textit{Color}, \textit{Texture}, \textit{Composition}), highlighting the fluidity of abstract art semantics.
	}
	\label{fig:umap_combined}
\end{figure}

\subsection{Framework of the Abstract Art Standard System}

To describe the non-representational features of abstract art, we propose a four-dimensional perceptual framework defining its core visual language---\textit{form}, \textit{color}, \textit{texture}, and \textit{composition} (Table~\ref{tab:framework}).
These dimensions are widely discussed in art-historical and formalist analyses as fundamental components of visual expression \cite{kress2020reading, kim2022formal, arnheim1954art, itten1970elements}.

This framework guides the annotation and analysis process in Abstract4D by linking artistic concepts with computational representations.
It also provides a structured foundation that can be extended in future work to incorporate additional perceptual or affective attributes.

\subsection{Data Collection and Processing}
\label{subsec:collection}

We compile artworks from a wide range of open and verified sources, including public museum archives, open-access art databases (e.g., WikiArt~\cite{wiki2025}, OmniArt~\cite{strezoski2018omniart}), and curated online galleries such as \textit{Saatchi Art} \cite{saatchiart} and \textit{Mei-shu Web} \cite{meishuwang}.
Each record contains metadata fields for artist name, artwork title, creation year, and nationality, which are normalized for consistency.
We remove low-quality, duplicated, or irrelevant entries using a two-stage process:
(1)~visual filtering to exclude blurry or incomplete artworks; and
(2)~semantic validation using CLIP~\cite{radford2021learning} similarity.
Each image's embedding is compared against two textual queries: \textit{``an abstract painting''} and \textit{``not an abstract painting''}.
Images with higher similarity to the latter are manually reviewed and discarded.
This hybrid verification ensures that all retained samples are genuinely abstract rather than figurative or decorative.

After filtering, the dataset contains approximately \textbf{120,000} valid images, covering over \textbf{14,044} artists across \textbf{150} countries and spanning artworks from the late 19th century to the present day.

\subsection{Prompt Annotation}

Each image in Abstract4D is paired with a prompt describing the four perceptual dimensions defined in Table~\ref{tab:framework}.
The annotations are produced through a hybrid human--VLM pipeline to balance scalability and reliability.

For each artwork, a vision--language model (GPT-4.5) first generates candidate descriptions conditioned on the image.
Two prompting strategies are used: a structured prompt that explicitly covers the four perceptual dimensions, and a free-form prompt that produces a more narrative description of the visual composition.

Human annotators review these candidates and synthesize the final prompt by selecting informative phrases and removing redundancy.
The model outputs serve only as proposals, while the final annotations are curated by human annotators.

\subsection{Dataset Overview and Statistics}

We analyze the Abstract4D dataset across multiple metadata and perceptual dimensions (Figure \ref{fig:dataset_overview}).
Overall, the dataset exhibits a rich temporal and geographic diversity, with the majority of artworks created between 1920 and 2020.
Figure \ref{fig:dataset_overview}(a) illustrates the artist distribution, highlighting the most prolific contributors.
Figure \ref{fig:dataset_overview}(b) shows the decade-wise growth, revealing a marked increase in artistic activity in the 21st century.
Figure \ref{fig:dataset_overview}(c) visualizes the national composition, indicating a globally distributed corpus with a few dominant contributing countries.

Table~\ref{tab:comparison} compares Abstract4D with prior art datasets.
While WikiArt and OmniArt provide large-scale style-labeled data, they contain limited purely abstract content and lack structured perceptual descriptions.
Abstract4D, by contrast, focuses exclusively on non-representational artworks and provides semantically rich multi-dimensional prompts aligned with visual perception.

\paragraph{Data Sources and Usage Policy}
The artworks in Abstract4D are collected from publicly accessible online sources, including open art databases, gallery websites, and online auction platforms.
The dataset is intended solely for academic research and non-commercial use.
Whenever available, we retain the original metadata associated with each artwork, including artist name, title, and source attribution.
The dataset does not claim ownership of the artworks, and the images remain the property of their respective copyright holders.

\begin{table}[t]
	\centering
	\caption{
		\textbf{Classification performance} of the CLIP base linear probe on the artist-level and nationality-level subsets of the Abstract4D dataset.
		All metrics are macro-averaged across classes.
	}
	\vspace{4pt}
	\small
	\resizebox{\columnwidth}{!}{%
		\begin{tabular}{lcccccc}
			\toprule
			\textbf{Task}  & \textbf{Precision} & \textbf{Recall} & \textbf{F1-score} & \textbf{Top1} & \textbf{Top5} & \textbf{Macro F1} \\
			\midrule
			Artist  & 0.680 & 0.496 & 0.487 & 0.7382 & 0.8328 & 0.7320 \\
			Nationality  & 0.272 & 0.196 & 0.164 & 0.1960 & 0.4383 & 0.1635 \\
			\bottomrule
		\end{tabular}%
	}
	\label{tab:clip_cls_metrics}
\end{table}

\section{Semantic Space Analysis}

This section analyzes the semantic organization of the Abstract4D corpus using the textual prompts associated with each artwork.
We encode prompts into a high-dimensional embedding space and explore its structure through dimensionality reduction, clustering, and metadata-aware analysis. This analysis aims to reveal how perceptual attributes of abstract artworks organize within a continuous semantic representation.

\subsection{Embedding Construction}
\label{subsec:embed}
Each prompt is encoded into a sentence-level embedding using a pretrained Sentence-BERT model \cite{reimers2019sentence} which is chosen for its ability to capture semantic similarity between textual descriptions while preserving contextual relationships between perceptual attributes.. These embeddings define a continuous high-dimensional semantic space in which proximity between vectors corresponds to perceptual affinity across artworks. Unless otherwise noted, all embeddings are $\ell_2$-normalized.

\subsection{Low-Dimensional Visualization}

We apply UMAP \cite{mcinnes2018umap} to project the high-dimensional prompt embeddings into two dimensions while preserving local neighborhood relationships and maintaining coarse global topology. UMAP was chosen for its ability to capture both fine-grained local structure and broader semantic patterns, which is essential for visualizing the continuous nature of perceptual attributes in abstract art. Figure~\ref{fig:umap_combined} (a) shows a uniform scatter plot without obvious density artifacts, suggesting that the prompts cover a broad spectrum of abstract-art semantics beyond simple style labels.

In this visualization, the continuous distribution of points indicates that the dataset's perceptual descriptions span diverse themes and relationships within the abstract art domain. The lack of clustering artifacts further confirms that the prompts do not correspond to discrete style categories but instead represent a range of perceptual features that abstract artworks convey.

\subsection{Semantic Clustering with Interpretive Labeling}

To identify semantic neighborhoods in the embedding space, we apply KMeans clustering \cite{lloyd1982least, pedregosa2011scikit} with $K{=}30$, chosen empirically to balance cluster interpretability and stability.
In parallel, we employ HDBSCAN \cite{campello2013density} to detect dense semantic communities and identify potential outliers.

Figure~\ref{fig:umap_combined}(b) visualizes the KMeans assignments on the UMAP projection.
Clusters exhibit gradual transitions rather than sharp boundaries, reflecting the continuous nature of perceptual attributes in abstract art.

For KMeans, the embeddings achieve a silhouette score of $\sim$0.20 and a Davies-Bouldin index of $\sim$1.7 \cite{davies2009cluster}.
These values indicate moderate but meaningful cluster separation, which is expected given that perceptual attributes in abstract art form overlapping semantic continua rather than discrete categories.
HDBSCAN further identifies a small set of outliers corresponding to rare or semantically mixed prompts.

\subsection{Cultural Patterns}
\label{subsec:nationality}

To explore cultural patterns in abstract visual language, we analyze the relationship between prompt-derived perceptual attributes and artist metadata.
For each perceptual dimension, we compute proxy-axis scores from the prompt embeddings and aggregate them by nationality.

\begin{figure}[t]
	\centering
	\includegraphics[width=\columnwidth]{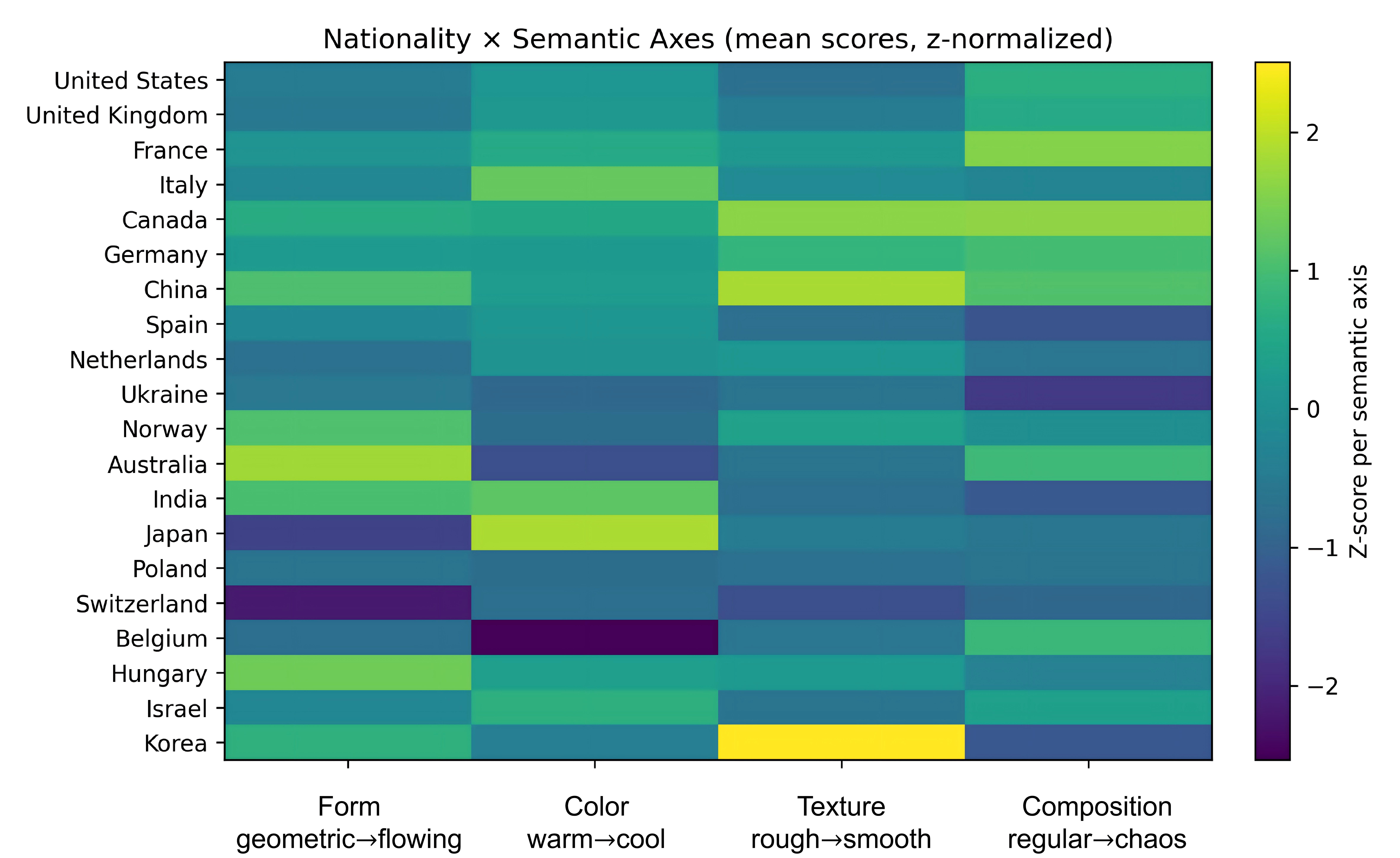}
	\caption{\textbf{Heatmap} showing the mean $z$-normalized scores of paintings across four semantic axes (Form, Color, Texture, Composition) grouped by nationality.}

	\label{fig:heatmap_nat}
\end{figure}

Figure~\ref{fig:heatmap_nat} shows the mean scores (z-normalized per axis) across the top-20 countries by sample size.
Differences appear across several perceptual dimensions, particularly along the form and composition axes.

While this analysis is exploratory, it illustrates how structured prompt annotations enable large-scale comparative studies of abstract art across cultural contexts.

\section{Baseline Experiments}

We conduct baseline experiments to demonstrate how the Abstract4D dataset can support computational analysis of abstract art.
The experiments include classification, cross-modal retrieval, and text-to-image generation tasks using pre-trained vision--language and diffusion models.
All experiments are conducted on NVIDIA RTX 4090 GPUs using the PyTorch framework.

\subsection{Classification}

To evaluate the discriminative capacity of pre-trained vision--language models on abstract artworks, we perform supervised classification on two subsets of Abstract4D labeled by \textit{artist identity} and \textit{nationality}. The dataset is randomly split into training, validation, and test sets with a ratio of 7:2:1.

We use the \textbf{CLIP base model} with a \textit{linear probe} configuration, where the image encoder is frozen and a linear classifier is trained on top of the extracted embeddings.
The classifier is trained for 30 epochs using the AdamW optimizer with standard data augmentations.

Models are evaluated using Precision, Recall, F1-score, Top-1, Top-5 accuracy, and Macro F1.
Confusion matrices are also analyzed to examine class separability.

\begin{figure}[t]
	\centering
	\includegraphics[width=\columnwidth]{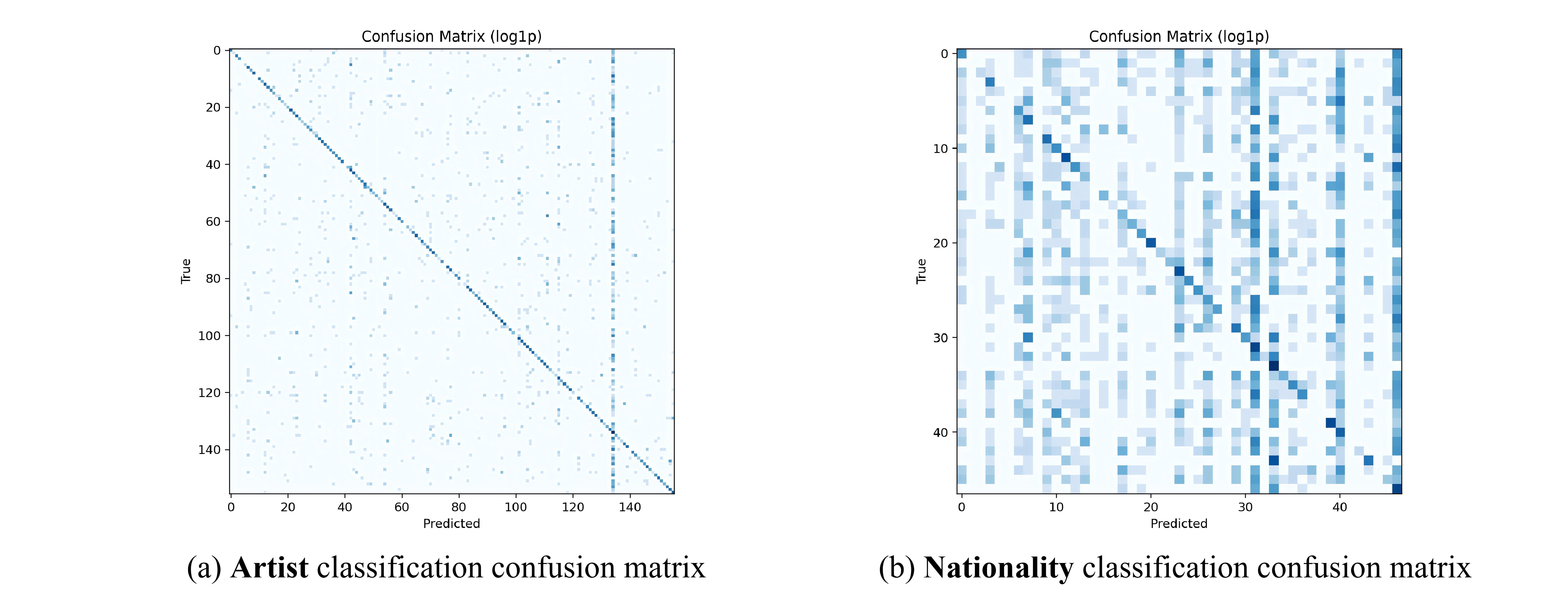}
	\caption{
		\textbf{Confusion matrices for CLIP linear probe classification.}
		(a) Artist classification shows clear diagonal dominance, indicating distinct individual styles.
		(b) Nationality classification exhibits dispersed patterns, reflecting weak cross-national regularities.
	}
	\label{fig:cls_confmat}
\end{figure}

As reported in Table~\ref{tab:clip_cls_metrics} and Figure~\ref{fig:cls_confmat}, the CLIP linear probe performs well on \textbf{artist classification} (Top-1 = 0.7382, Macro F1 = 0.7320), but performs significantly worse on \textbf{nationality classification} (Top-1 = 0.1960, Macro F1 = 0.1635).
This result suggests that artist-specific visual patterns are more distinguishable than nationality-level characteristics in abstract artworks.

\begin{figure*}[t]
	\centering
	\includegraphics[width=\linewidth]{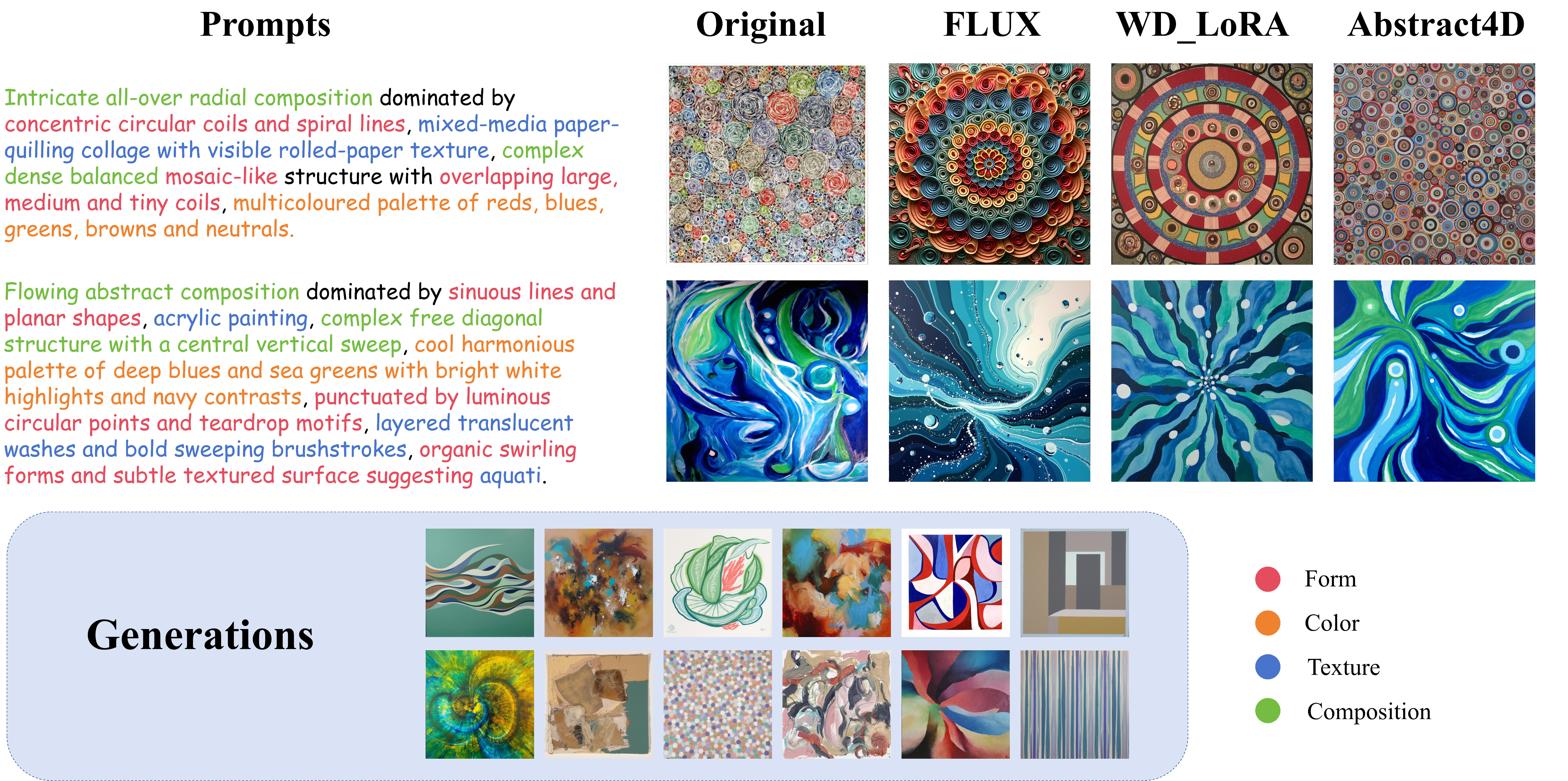}
	\caption{
		\textbf{Qualitative comparison of \textit{Flux} generative results.}
		The main grid (upper area) shows three representative prompts (left) with their reference paintings
		and generations from \textbf{Flux (no finetune)}, \textbf{Flux + WD-LoRA}, and \textbf{Flux + Abstract4D-LoRA}.
		The Abstract4D-LoRA model produces outputs better meet the four-dimensional description requirements and closely align with the reference images, demonstrating improved compositional balance, coherent chromatic relationships, and painterly texture realism.
		The blue panel at the bottom-left (\textit{Generations}) displays additional artworks freely generated by the Abstract4D-LoRA model, illustrating its creative diversity and fidelity to the visual language of abstract art.}
	\label{fig:generation}
\end{figure*}

\subsection{Retrieval}

We evaluate cross-modal retrieval on the Abstract4D dataset, where each image corresponds to a unique textual description.
The model backbone is CLIP ViT-B/32.
The \textit{zero-shot} baseline corresponds to the pretrained CLIP evaluated directly on Abstract4D.
The \textit{fine-tuned} model is trained with a contrastive InfoNCE loss on 80k image--text pairs, using AdamW ($\text{lr}=1\times10^{-5}$, $\text{wd}=0.01$), batch size $64$ (effective $256$ with accumulation), input size $224$, and trained for 12~epochs.
We report Recall@K, Median Rank (MedR), and nDCG@10 for both text$\to$image (T$\to$I) and image$\to$text (I$\to$T) directions.

Table~\ref{tab:retrieval_main} compares the zero-shot model and the final fine-tuned model (Epoch 12).
Fine-tuning yields consistent and substantial gains, improving mean R@1 from $0.717$ to $0.895$ and achieving perfect median rank ($\text{MedR}=1$) with nDCG@10 above $0.92$.
The improvement is symmetric between the two retrieval directions, indicating that both encoders benefit equally from domain adaptation.

The zero-shot model already achieves moderate alignment (mean~0.72), and performance steadily improves across epochs, converging near 0.89.
This continuous upward trend confirms that even lightweight training suffices for CLIP to internalize the abstract visual--textual correspondences present in Abstract4D.

Fine-tuning substantially enhances CLIP's capability to bridge abstract textual and visual semantics.
The balanced gains across both retrieval directions suggest a coherent shared embedding space.
These results confirm that Abstract4D provides sufficient signal to teach a pretrained vision--language model new aesthetic and perceptual associations, even under limited training epochs.

\begin{table}
	\centering
	\caption{
		Text--image \textbf{retrieval results} on the Abstract4D validation and test sets.
		Fine-tuning substantially improves cross-modal alignment over the zero-shot CLIP baseline.
	}
	\vspace{4pt}
	\small
	\begin{tabular}{lcccc}
		\toprule
		\textbf{Model} & \textbf{Dir.} & \textbf{R@1} & \textbf{MedR} & \textbf{nDCG@10} \\
		\midrule
		Zero-shot CLIP & T$\to$I & 0.7035 & 5.0 & 0.7363 \\
		& I$\to$T & 0.7308 & 4.0 & 0.7590 \\
		\midrule
		Fine-tuned CLIP & T$\to$I & \textbf{0.8934} & \textbf{1.0} & \textbf{0.9226} \\
		& I$\to$T & \textbf{0.8956} & \textbf{1.0} & \textbf{0.9240} \\
		\bottomrule
	\end{tabular}
	\label{tab:retrieval_main}
\end{table}

\subsection{Text-to-Image Generation}

To evaluate how structured perceptual prompts influence generative modeling, we fine-tune the diffusion model \textbf{Flux} \cite{labs2025flux1kontextflowmatching, flux2024} using LoRA \cite{hu2022lora}.
Figure~\ref{fig:generation} presents qualitative comparisons of generated abstract artworks.

We randomly sample 1,000 paintings from Abstract4D, each paired with its four-dimensional prompt.
The model is fine-tuned with a learning rate of $1\times10^{-4}$, batch size 64, and 10k training steps using the AdamW optimizer.
The base Flux weights remain frozen, and only LoRA layers are updated during training.

For comparison, we train another LoRA model using pseudo-prompts generated by \textbf{WD-Tagger} \cite{wdtagger2025}, a tagging model that produces descriptive tags for artworks.
All models are evaluated under identical inference settings (CFG=7.5, 50 DDIM steps, $512{\times}512$ resolution).

As illustrated in Figure~\ref{fig:generation}, the \textbf{Abstract4D-LoRA} model generates images that better reflect the perceptual relationships specified in the prompts, particularly in terms of compositional balance, color structure, and texture consistency.
These qualitative results suggest that structured perceptual descriptions can help guide generative models toward more coherent abstract visual compositions.

\section{Conclusion}

In this work, we introduced Abstract4D, the largest dataset of abstract paintings, providing rich multi-dimensional annotations across four perceptual dimensions: form, color, texture, and composition. Through a hybrid human-AI annotation pipeline, we have built a robust foundation for AI to understand the visual language of abstract art. Our experiments, including classification, retrieval, and generation tasks, showcase significant advancements in AI's ability to internalize and represent abstract visual language, especially with the help of fine-tuning. The results demonstrate that AI models can be effectively trained to recognize and generate abstract art that adheres to its inherent visual grammar. Through Abstract4D, we provide not only a new dataset but also a comprehensive framework for bridging the gap between artistic understanding and computational models, marking a step forward in AI's engagement with abstract art.

\bibliographystyle{unsrtnat}
\bibliography{references}

\end{document}